\documentclass[10pt,twocolumn,letterpaper]{article}

\usepackage{cvpr}
\usepackage{times}
\usepackage{epsfig}
\usepackage{amsmath}
\usepackage{amssymb}
\usepackage{helvet} 
\usepackage{courier}  
\usepackage[hyphens]{url}  
\usepackage{graphicx} 
\usepackage{graphics} 
\usepackage{mathptmx} 
\usepackage{amsmath,amssymb} 
\usepackage{color}
\usepackage{multirow}
\usepackage{longtable}
\usepackage{rotating}
\usepackage{array}
\usepackage{float}
\usepackage{subfigure}
\usepackage{algorithm}
\usepackage{algorithmic}

\usepackage{xcolor}

\usepackage[breaklinks=true,bookmarks=false]{hyperref}

\cvprfinalcopy 

\def\cvprPaperID{****} 

\begin{document}

\title{Domain Balancing: Face Recognition on Long-Tailed Domains}


\author{Dong Cao$^{1,2}$\thanks{Equally-contributed}\quad Xiangyu Zhu$^{1,2}$\footnotemark[1]\quad Xingyu Huang$^{3}$ \quad Jianzhu Guo$^{1,2}$\quad Zhen Lei$^{1,2}$\thanks{Corresponding author} \\
$^{1}$CBSR $\&$ NLPR, Institute of Automation, Chinese Academy of Sciences, Beijing, China\\
$^{2}$School of Artificial Intelligence, University of Chinese Academy of Sciences, Beijing 100049, China\\
$^{3}$Tianjin University\\
{\tt\small \{dong.cao,xiangyu.zhu,jianzhu.guo,zlei\}@nlpr.ia.ac.cn, xingyu.huang@tju.edu.cn}
}
\maketitle

\thispagestyle{empty}
\pagestyle{empty}
\begin{abstract}
Long-tailed problem has been an important topic in face recognition
task. However, existing methods only concentrate on the long-tailed distribution of classes. Differently, we devote to the long-tailed domain distribution problem, which refers to the fact that a small number of domains frequently appear while other domains far less existing.
The key challenge of the problem is that domain labels are too complicated (related to race, age, pose, illumination, etc.) and  inaccessible in real applications. In this paper, we  propose a novel Domain Balancing (DB) mechanism to handle this problem. Specifically, we first propose a Domain Frequency Indicator (DFI) to judge whether a sample is from head domains or tail domains. Secondly, we formulate a light-weighted Residual Balancing Mapping (RBM) block to balance the domain distribution by adjusting the network according to DFI.
Finally, we propose a Domain Balancing Margin (DBM) in the loss function to further optimize the feature space of the tail domains to improve generalization. Extensive analysis and experiments on several face recognition benchmarks demonstrate that the proposed method  effectively enhances the generalization capacities and achieves superior performance.
\end{abstract}

\section{Introduction}
Feature descriptor is of crucial importance to the performance of face recognition, where the training and testing images are drawn from different identities and the distance metric is directly acted on the features to determine whether they belong to the same identity or not.
Recent years have witnessed remarkable progresses in face recognition, with a variety of approaches proposed in the literatures and applied in real applications \cite{liu2017sphereface,wang2018cosface,deng2019arcface,guo2020learning,guo2018face,zhu2019large}.
Although yielding excellent success, face recognition often suffers from poor generalization, i.e., the learned features only work well on the domain the same as the training set and perform poorly on the unseen domains.
This is one of the most critical issues for face recognition in the wild, partially due to the non-negligible domain shift from the training set to the deployment environment.

\begin{figure}[tbp]
\includegraphics[scale=0.4]{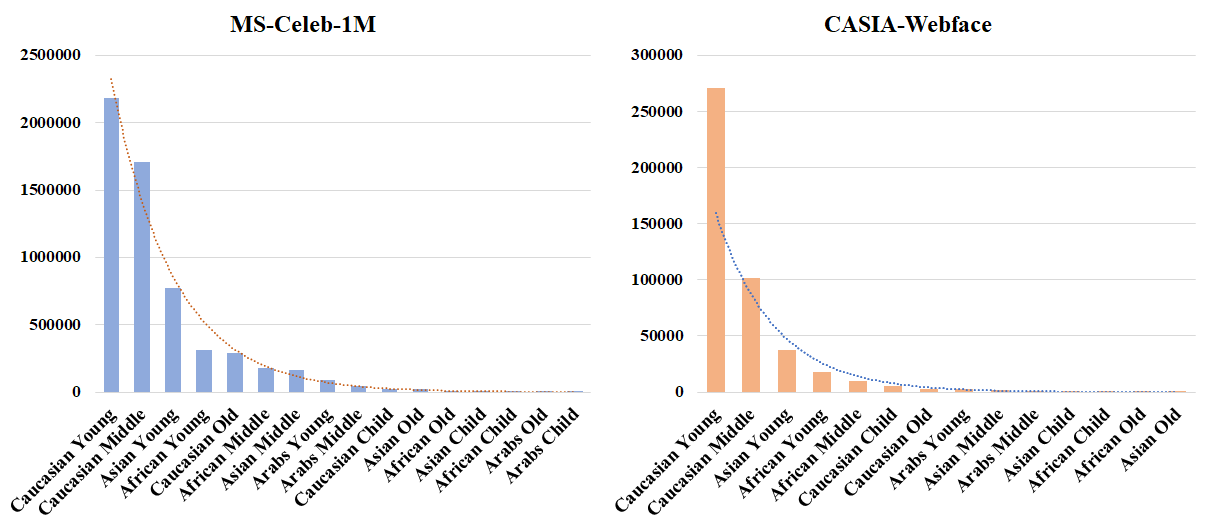}
\caption{\label{img:summ-dataset} The long-tailed domain distribution demarcated by the mixed attributions (e.g., race and age) in the MS-Celeb-1M \cite{guo2016ms} and CASIA-Webface \cite{yi2014learning}. Number of classes per domain falls drastically, and only few domains have abundant classes. (Baidu API \cite{baidu} is used to estimate the race and age)}
\end{figure}

Real-world visual data inherently follows a long-tailed distribution, where only a limited number of classes appear frequently, and  most of the others remain relatively rare.
In this paper, we aim to investigate the long-tailed domain distribution and balance it to improve the generalization capacity of deep models. However, different from the long-tailed problem in classes, domain labels are inaccessible in most of applications. Specifically, domain is an abstract attribute related to many aspects,
e.g., age (baby, child, young man, aged, etc), race (caucasian, indian, asian, african, etc.), expression (happy, angry, surprise, etc.), pose (front, profile, etc.), etc.
As a result, the domain information is hard to label or even describe. Without the domain label, it is difficult to judge whether a sample belongs to the head domains or the tail domains, making existing methods inapplicable.
Figure~\ref{img:summ-dataset} illustrates a possible  partition by the mixed attributions (e.g., race and age).

We formally study this long-tailed domain distribution problem arising in real-world data settings.
Empirically, the feature learning process will be significantly dominated by those few head domains while ignoring many other tail domains, which increases the recognition difficulty in the tail domains.
Such undesirable bias property poses a significant challenge for face recognition systems, which are not restricted to any specific domain. Therefore, it is necessary to enhance the face recognition performance regardless of domains.
An intuitive method to handle the long-tailed problem is over-sampling and under-sampling samples on the tail and the head, respectively \cite{he2009learning,shen2016relay,zhong2016towards}. However, it does not work well on domain balancing since the the ground-truth domain distribution is inaccessible. To overcome this drawback, we propose a Domain Balancing (DB) mechanism to balance the long-tailed domain distribution.

\begin{figure}[tbp]
\centering
\includegraphics[scale=0.18]{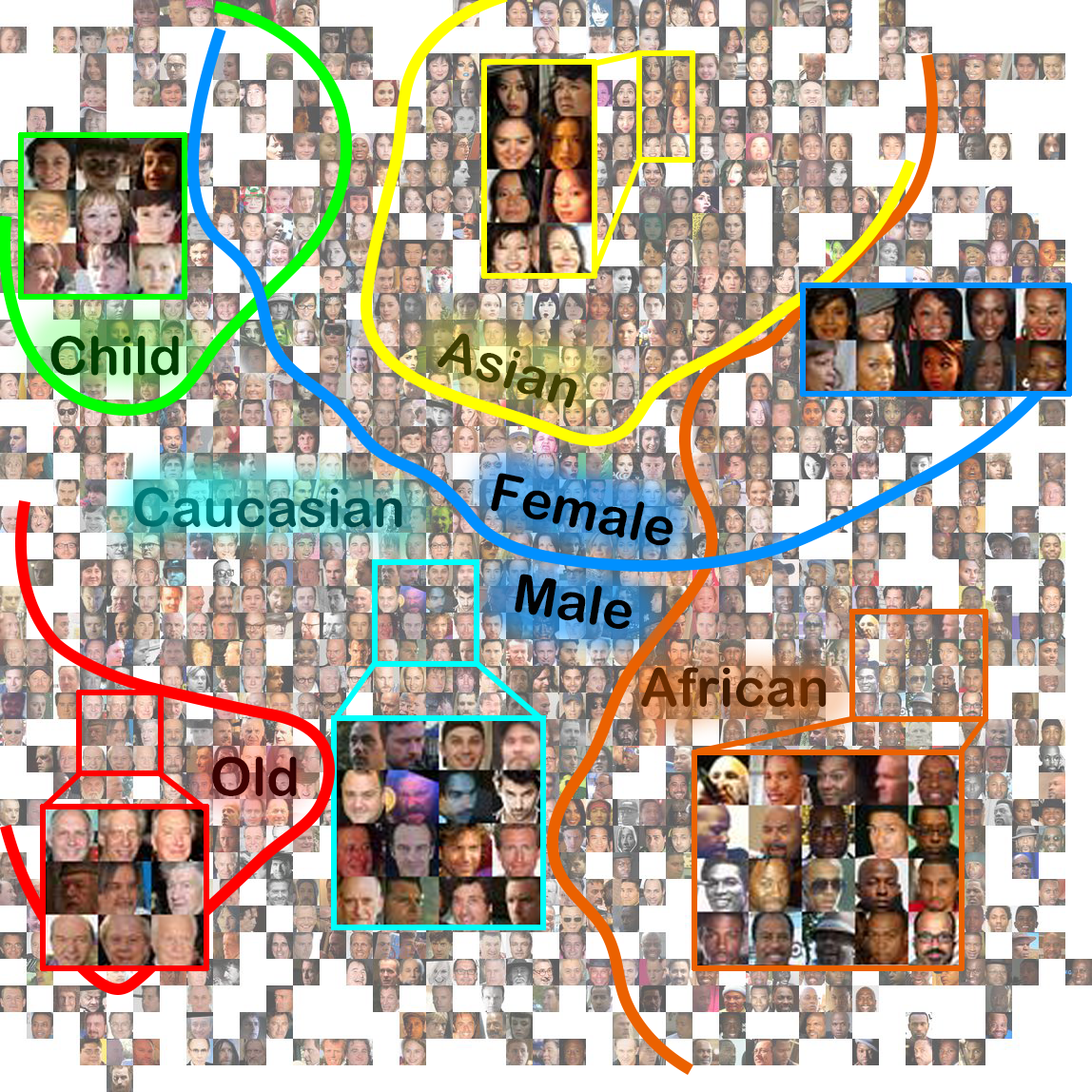}
\caption{\label{img:distribution}The face features are trivially grouped together according to different attributions, visualized by t-SNE \cite{maaten2008visualizing}}
\end{figure}

\begin{figure*}[tbp]
\centering
\includegraphics[scale=0.44]{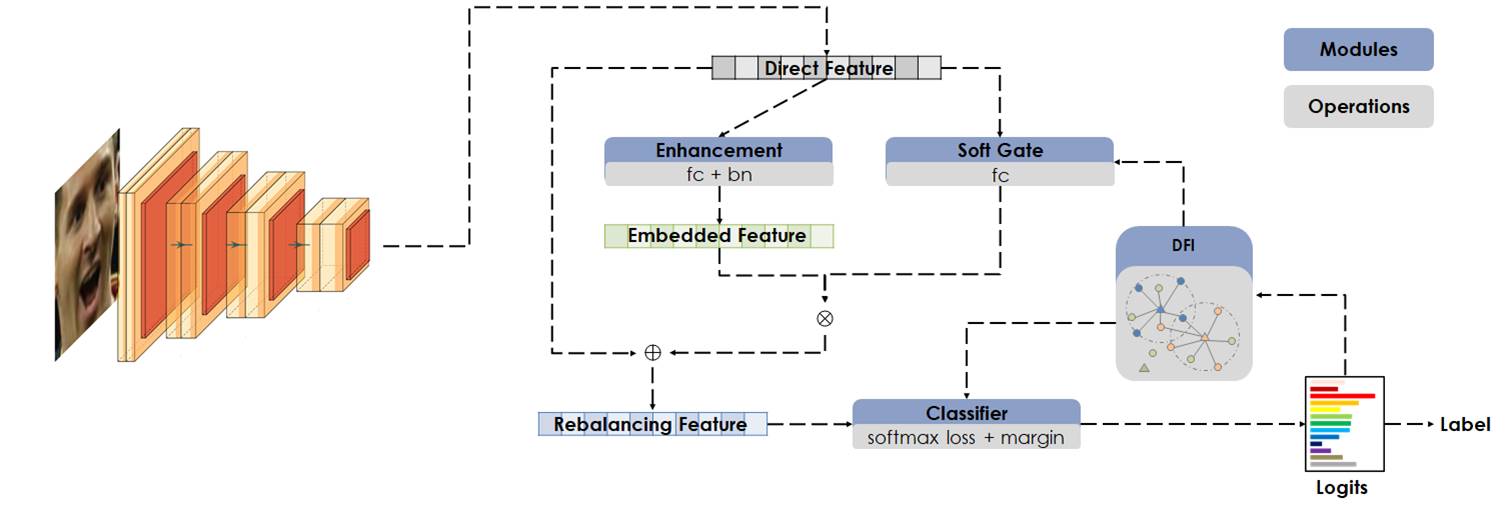}
\caption{\label{img:framework} There are three main modules: DFI, RBM and DBM. The DFI indicates the local distances within a local region.
The RBM harmonizes the representation ability in the network architecture, while the DBM  balances the contribution in the loss.}
\end{figure*}
Firstly, since the ground truth domain distributions are inaccessible without domain labels, for each sample, we should predict where the belonged domain locates on the distribution. To this end, we point out that the domain frequency can be instructed by the inter-class compactness. In the holistic feature space, the classes with similar attributes tend to group, forming a specific domain as shown in Figure~\ref{img:distribution}.  Besides,
in the feature space, the compactness is not everywhere equal.
Take the domains in Figure~\ref{img:summ-dataset} as an example, we find
the compact regions tend to belong to the head domains (e.g., caucasian male), and the sparse regions tend to belong to the tail domains (e.g., children, african female). The detailed analysis will be shown in Section~\ref{DFI}.
Motivated by these observations, we propose to utilize the inter-class compactness which is the local distances within a local region as the Domain Frequency Indicator (DFI).
Secondly, considering the samples belong to the same domain share some appearance consistency, we design a novel module called Residual Balancing Mapping (RBM) block, which can adaptively change the network based on DFI to find the best network to adjust each domain.
The block consists of two components: a domain enhancement branch and a soft gate.
The domain enhancement branch aims to adjust the network to each domain through enhancement residual and the soft gate attaches a harmonizing parameter to the residual to control the amount of residual according to the domain frequency.
Thirdly,
in the loss function, we propose a Domain Balancing Margin (DBM) to
adaptively modify the margin according to the DFI for each class, so that the loss produced by the  tail domain classes  can be relatively up-weighted. The framework is shown in Figure~\ref{img:framework}.

The major contributions can be summarized as follows:
\begin{itemize}
  \item We highlight the challenging long-tailed domain problem,  where we must balance the domain distribution without any domain annotation.
  \item We propose a Domain Balancing (DB) mechanism to solve the long-tailed domain distribution problem. The DB can automatically evaluate the  domain frequency of each class with a Domain Frequency Indicator (DFI) and adapt the network and loss function with Residual Balancing Mapping (RBM) and Domain Balancing Margin (DBM), respectively.
  \item We evaluate our method on several large-scale face datasets. Experimental results show that
        the proposed Domain Balancing can  efficiently mitigate the long-tailed domain distribution problem  and outperforms the state-of-the-art approaches.
\end{itemize}

\section{Related Works}
$\mathbf{Softmax\ \ based\ \ Face\ \ Recognition.}$
Deep convolutional neural networks (CNNs) \cite{Cun1995Convolutional} have achieved impressive success in
face recognition.
The current prevailing softmax loss considers the training process as a
N-way classification problem.
Sun et al. \cite{sun2014deep} propose the DeepID for face verification.
In the training process, for each sample, the extracted feature is taken to calculate the dot products with all the class-specific weights.
Wen et al. \cite{wen2016discriminative} propose a new center loss penalizing the distances between the features and their corresponding class centers.
Wang et al. \cite{wang2017normface} study the effect of normalization during training and show that optimizing cosine similarity (cosine-based softmax loss)
instead of inner-product improves the performance.
Recently, a variety of margin based softmax losses \cite{liu2017sphereface,wang2018cosface,deng2019arcface} have achieved the state-of-the-art performances.
SphereFace \cite{liu2017sphereface} adds an extra angular margin to attain shaper decision boundary of the original softmax loss. It concentrates the features in a sphere mainfold. CosFace \cite{wang2018cosface} shares a similar idea which encourages the intra-compactness in the cosine manifold.
Another effort ArcFace \cite{deng2019arcface} uses an additive angular margin, leading to similar effect. However, these efforts only consider the intra-compactness. RegularFace \cite{zhao2019regularface} proposes an exclusive regularization to focus on the inter-separability.
These methods mainly devote to enlarge the inter-differences and reduce the intra-variations.
Despite their excellent performance on face recognition, they rely more on the large and balanced datasets and often suffer performance degradation when facing with the long-tailed data.


$\mathbf{Long-tailed\ \ Learning}$
Long-tailed distribution of data has been well studied in \cite{zhang2017range,liu2019large}.
Most existing methods define the long-tailed distribution in term of the size of each class.
A widespread method is to resample and rebalance training data, either by under-sampling examples from the head data \cite{he2009learning}, or over-sampling samples from the rare data more frequently \cite{shen2016relay,zhong2016towards}. The former generally loses critical information in the head sets, whereas the latter generates redundancy and may easily encounter the  problem of over-fitting to the rare classes.
Some recent strategies include hard negative mining \cite{dong2017class,lin2017focal}, metric learning \cite{huang2016learning,oh2016deep} and meta learning \cite{ha2016hypernetworks,wang2017learning}. The range loss \cite{zhang2017range} proposes an extra range constraint jointly with the softmax loss.
It reduces the $k$ greatest intra-class ranges and enlarges the shortest inter-class distance within one batch. The focal loss \cite{lin2017focal} employs an online version of
hard  negative mining. Liu et al. \cite{liu2019large} investigate the long-tailed problem in the open set. Its so-called dynamic meta-embedding uses an associated memory to enhance the representation.
Adaptiveface \cite{liu2019adaptiveface} analyzes the difference between rich and poor classes and proposes the adaptive margin softmax to dynamically modify the margins for different classes.
Although the long-tailed problem has been well studied, they are mainly based on the category
frequency distribution.
None of previous works consider the similar problem in domain. One possible reason may be due to the ambiguous domain partition as discussed above.
In fact, the domains may not even have explicit semantics, i.e., they are actually data-driven.

In contrast, our method focuses on the long-tailed domain, which is more in line with the real-world application. The proposed method balances the contribution of domains on the basis of their frequency distribution, so that it can improve the poor generalization well.

\begin{figure}[tbp]
\centering
\includegraphics[scale=0.32]{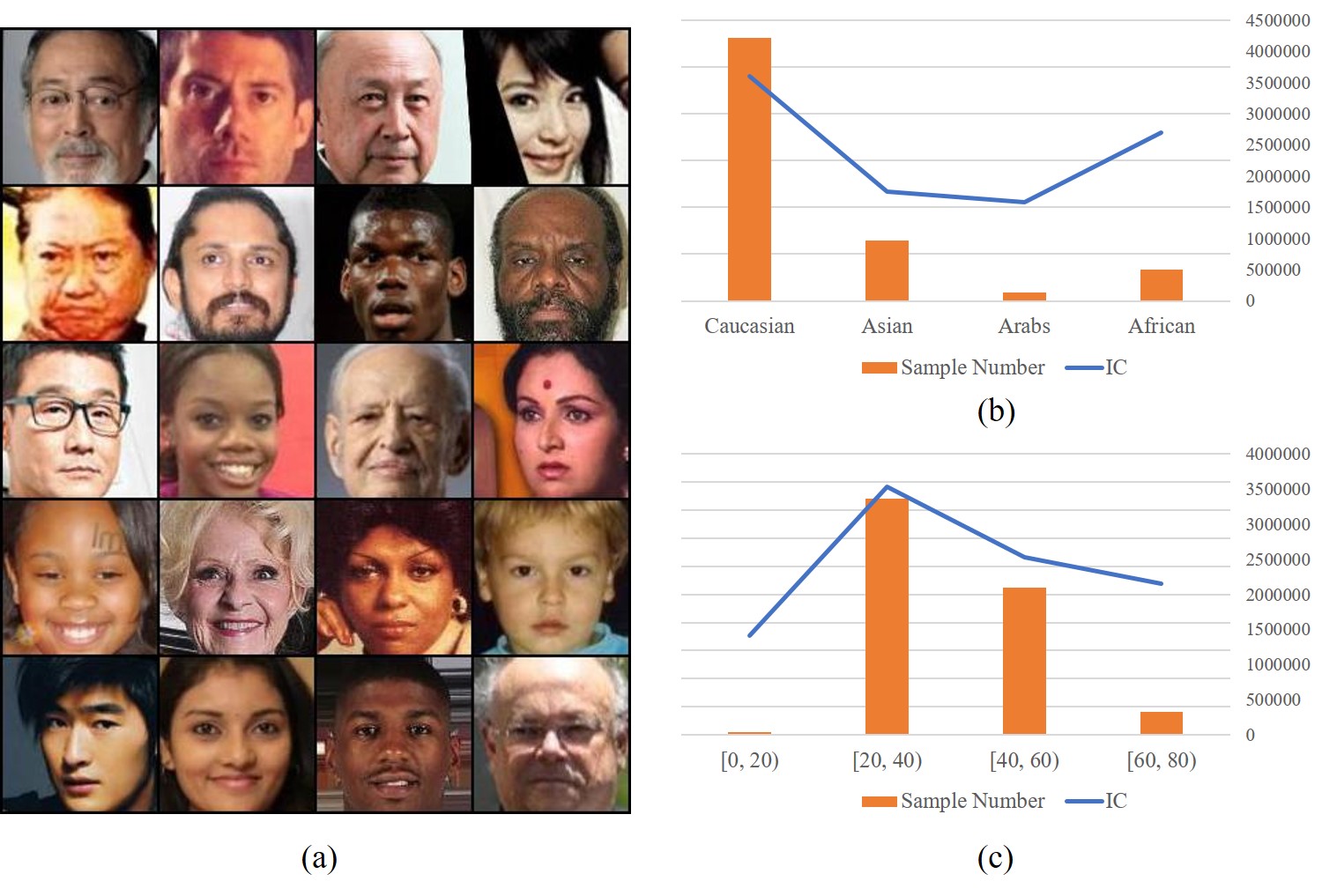}
\caption{\label{img:beta}(a) Identities with small inter-class compactness value in the MS-Celeb-1M. (b) The inter-class compactness vs. race distribution. (c) The inter-class compactness vs. age distribution.}
\end{figure}

\section{Domain Balancing}
We propose to balance the samples from  different domains without any domain annotation.
Domain Balancing (DB) mechanism has three components: Domain Frequency Indicator (DFI) to evaluate the domain frequency, the Residual Balancing Mapping (RBM) to adjust the network and the Domain Balancing Margin (DBM) to adjust the loss functions according to domain distribution.

\subsection{Domain Frequency Indicator} \label{DFI}
To handle the long-tailed domain distribution problem, we first need to know whether a sample is from a head domain or from a tail domain. We introduce a Domain Frequency Indicator (DFI) based on the inter-class compactness. Inter-class compactness function of a given class is formulated as:
\begin{equation}
\begin{aligned}
\label{eq:compactness1}
IC(w)=log\sum_{k=1}^{K}e^{s  \cdot cos(w_{k},w)}
\end{aligned}
\end{equation}
where $w$ is the prototype of one class in the classification layer and $k$ is the $k$-th nearest class,
where the distance of two classes $i,j$ is formulated as $cos(w_{i},w_{j})$.
The high frequency domain, i.e., head domain, usually corresponds to
a large $IC(w)$, and vice versa.
Then we define the Domain Frequency Indicator as:
\begin{equation}
\begin{aligned}
\label{eq:compactness2}
DFI=\frac{\epsilon}{IC(w)}
\end{aligned}
\end{equation}
which is inversely proportional to the inter-class compactness $IC(w)$ and $\epsilon$ is a
constant value. Ideally, if the classes are uniformly distributed, each class
will have the same DFI. Otherwise, the classes with larger DFI
are more likely to come from a tail domain and should be relatively  up-weighted.
As shown in Figure~\ref{img:beta}, the identities with larger DFI values usually come from
Africa, children or the aged, which are highly related with the tail domains.
\begin{figure}[tbp]
\centering
\includegraphics[scale=0.4]{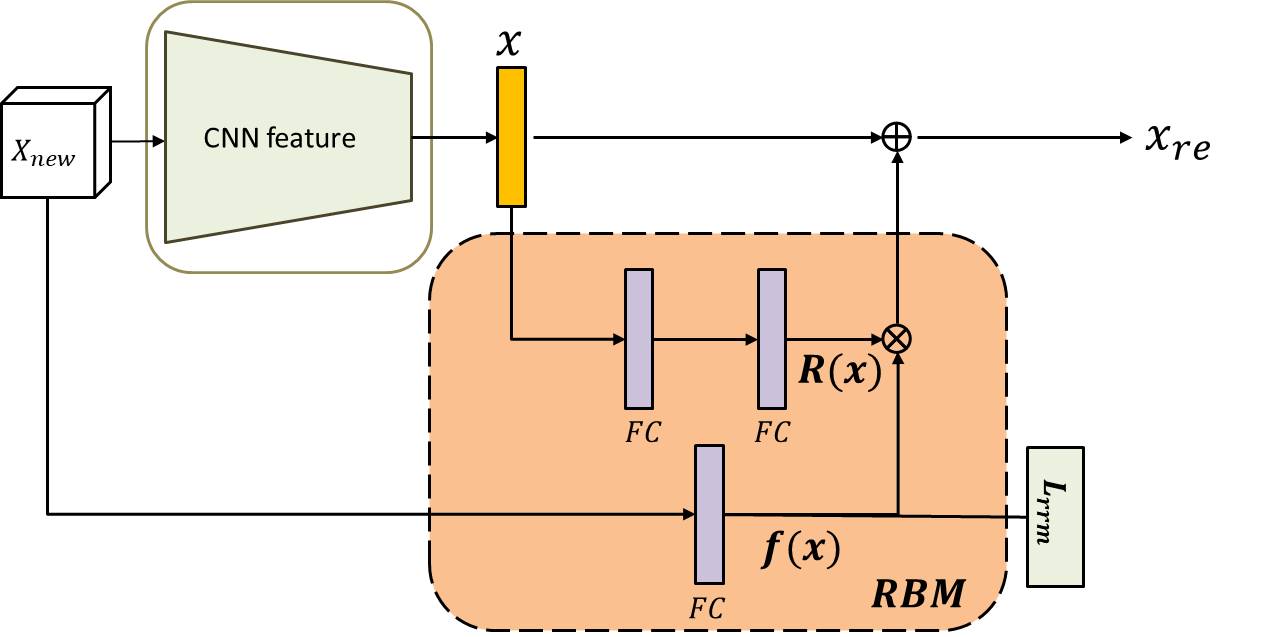}
\caption{\label{img:RRM}The Residual Balancing Module is designed with light-weighted structure and
it can be easily attached to existing network architecture. The block dynamically enhances the feature according
to DFI.
}
\end{figure}

\subsection{Residual Balancing Module}
In real-world application, face recognition accuracy depends heavily on the quality of the top-level feature $x$. The goal in this section is to design
a light-weight solution to adjust the network to extract domain specific features according to the domain distribution. Our Residual Balancing Module (RBM) combines the top-level image feature and a residual feature, using DFI to harmonize the magnitude.

Even though big training data facilitates the feature discriminative power, the head domains  dominate the learning process and the model lacks  adequate supervised updates from the tail classes. We hope to learn a harmonizing module through a mapping function $M_{re}(.)$ to adjust the features for samples of different domain frequency to mitigate the domain imbalance problem.
We formulate $M_{re}$ as a sum of the original feature $x$ and residual acquired by a  feature enhancement module $R(x)$ weighted by $f(x)$. We denote the resulting feature as $x_{re}$ and the RRM can be formulated as:
\begin{equation}
\begin{aligned}
\label{eq:residual}
x_{re}\ \ &=\ \ M_{re}(x)\\
&=\ \ x+f(x)\cdot R(x)
\end{aligned}
\end{equation}
where $x$ is the top-level feature, $f(x)$ is a soft gate depending on the DFI.
When DFI is large, the input feature probably belongs to a tail class, and
a large enhancement is assigned to the residual. Otherwise, the enhancement is trivial.
The magnitude of residual is thus inversely proportional to the domain frequency.
The combination of the soft gate and and the residual can be regarded as a harmonizing mechanism that adopts domain distribution information to control the magnitude to be passed to the next layer.

We now describe the implementation of the two components:
The first component is the residual $R(x)$, which is implemented by a light-weighted full-connected layer. It consists of two full-connected layers and a batch norm layer shown in Figure~\ref{img:RRM}. The second component is the soft gate coefficient DFI, which is learned  from the feature $x$ and supervised by the DFI. For simplicity, the linear regression is employed by the L2 loss:
\begin{equation}
\begin{aligned}
\label{eq:softmaxloss}
L_{rrm}=\|f(x)-DFI(x)\|_{2}^{2}
\end{aligned}
\end{equation}
where $DFI(x)$ is defined  in Eq.~\ref{eq:compactness2} from the last iteration. $f(x)$ is a mapping function devoting to associate the representation $x$ and DFI.
\begin{figure}[tbp]
\centering
\includegraphics[scale=0.175]{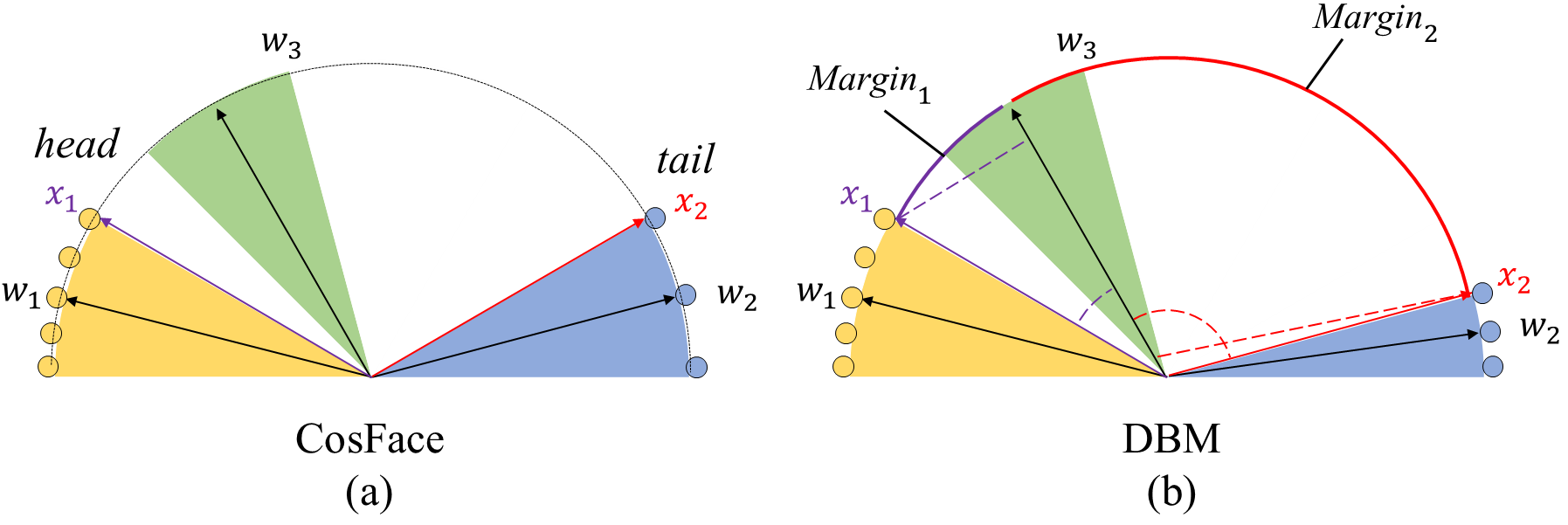}
\caption{\label{img:margin}Geometrical interpretation of DBM from the feature perspective.
Different color areas indicate feature space from distinct classes. Yellow area represents the head-domain class  $C_{1}$ and
blue area represents the tailed-domain class $C_{2}$.
(a) CosFace assigns an uniform margin for all the classes. The sparse inter-class distribution in the tail domains makes the decision boundary easy to satisfy. (b) DBM assigns margin according to the inter-class compactness adaptively.
}
\end{figure}
\subsection{Domain Balancing Margin}
We propose a domain-aware loss by Domain Balancing Margin (DBM) to adaptively strengthen the classes in the tail domains. Specifically, we formulate the DBM loss by embedding the DFI into CosFace as:
\begin{equation}
\begin{aligned}
\label{eq:softmaxloss}
L_{dbm}=-logP_{i,y_{i}}=-log\frac{e^{s(cos\theta_{i,y_{i}}-\beta_{y_{i}} \cdot m)}}{e^{s( cos\theta_{i,y_{i}}-\beta_{y_{i}} \cdot m)}+\sum^{C}_{k\neq y_{i}}e^{s\cdot cos\theta_{i,k}}}
\end{aligned}
\end{equation}
where $\beta_{y_{i}}=DFI_{y_{i}}$ and $m$ is a fixed parameter as defined in CosFace. Figure~\ref{img:margin} visualizes the phenomenon through a trinary classification. The main difference between DBM and CosFace is that our margin is dynamic and feature compactness related.
For the CosFace, the decision boundary assigns the same margin without considering the feature compactness. It cannot efficiently compact the feature space of the tailed-domain class $C_{2}$ since the sparse inter-class distribution makes the decision boundary easy to satisfy. The termination of optimization is so early, leading to poor generalization.
In contrast, our DBM drives adaptive decision boundary in terms of the inter-compactness, where $margin_{2}$ (tailed-domain margin) should be much larger than $margin_{1}$ (head-domain margin). Consequently, both the inter-separability and  the intra-compactness can be guaranteed.

We combine the mentioned $L_{dbm}$ and $L_{rrm}$ by a parameter $\lambda$. The final loss function can be formulated as:
\begin{equation}
\begin{aligned}
\label{eq:softmaxloss}
L=L_{dbm}+ \lambda L_{rrm}
\end{aligned}
\end{equation}

%
%
%
%
%
%
%
%

\section{Experiments}

\subsection{Datasets}
$\mathbf{Training \ \ Set.}$
We employ CASIA-Webface \cite{yi2014learning} and MS-Celeb-1M \cite{guo2016ms} as our training sets.
CASIA-WebFace is collected from the web. The face images are collected from various professions and suffer from large variations in illumination, age and pose.
MS-Celeb-1M is one of the largest real-world face datasets containing 98,685 celebrities and 10 million images. Considering the amount of noise, we use a refined version called MS1MV2 \cite{deng2019arcface} where a lot of manual annotations are employed to guarantee the quality of the dataset.

$\mathbf{Testing\ \ Set.}$
During testing, we firstly explore databases (RFW \cite{wang2018racial}, AFW \cite{chen2014cross}) with obvious domain bias to check the improvement. RFW is a popular benchmark for racial bias testing, which contains four  subsets, Caucasian, Asian, India and African.
Moreover, we collect a new dataset from CACD \cite{chen2014cross}, called Age Face in-the-Wild (AFW). We construct three testing subsets, Young (14-30 years old), Middle-aged (31-60 years old) and Aged (60-90 years old). Each subset contains 3,000 positive pairs and 3,000 negative pairs respectively. Besides, we further report the performance on several widely used benchmarks
including LFW \cite{huang2008labeled}, CALFW \cite{zheng2017cross}, CPLFW \cite{zheng2018cross} and AgeDB \cite{moschoglou2017agedb}.
LFW contains color face images from 5,749 different persons in the web. We verify the performance on 6,000 image pairs following the standard protocol of unrestricted with labeled outside data.
CALFW is collected with obvious age gap to add aging process intra-variance on the Internet.
Similarly, CPLFW is collected in terms of pose difference.
AgeDB contains face images from 3 to 101 years old. We use the most challenging subset
AgeDB-30 in the following experiments.
We also extensively evaluate our proposed method on large-scale face dataset, MegaFace \cite{kemelmacher2016megaface}.
MegaFace is one of the most challenging benchmark for large scale face identification and
verification.
The gallery set in MegaFace includes 1M samples from 690K individuals and the probe set
contains more than 100K images of 530 different individuals from FaceScrub \cite{ng2014data}.
Table~\ref{tab:dataset} shows the detailed information of the involved datasets.

\begin{table}[h]
\caption{Statistics of face datasets for training and testing. (P) and (G) indicates the
probe and gallery set respectively.}
\label{tab:dataset}
\begin{center}
\begin{tabular}{|c||c|c|c|}
\hline
 & Dataset &  Identities&  Images\\
\hline
\multirow{2}*{Training} &CASIA \cite{yi2014learning} & 10K & 0.5M \\
      &MS1MV2 \cite{deng2019arcface} & 85K & 5.8M \\
\hline
\multirow{7}*{Testing}& LFW \cite{huang2008labeled} & 5749 & 13,233 \\
  & CPLFW \cite{zheng2018cross} & 5,749 & 12,174 \\
  & CALFW \cite{zheng2017cross} & 5,749 & 11,652 \\
  & AgeDB \cite{moschoglou2017agedb} & 568 & 16,488 \\
  & RFW \cite{wang2018racial} & 11,430 & 40,607 \\
  & CACD \cite{chen2014cross}& 2,000 & 160,000 \\
  & MegaFace \cite{kemelmacher2016megaface} & 530 (P)& 1M(G)\\
\hline
\end{tabular}
\end{center}
\end{table}
\subsection{Experimental Settings}
For data prepossessing, the face images are resized to  $112\times 112$ by employing five facial points, and each pixel in RGB images is normalized by subtracting 127.5 and dividing by 128.
For all the training data, only horizontal flipping is used for data augmentation.
For the embedding neural network, we employ the widely used CNNs architectures, ResNet18 and ResNet50 \cite{he2016identity}. They both contain four residual blocks and finally produce a 512-dimension feature.

\begin{table*}[!htbp]
\caption{Face verification results ($\%$) with different strategies. (CASIA-Webface, ResNet18, RBM (w/o sg) refers to RBM without the soft gate, i.e., $f(x)=1$.)}
\label{tab:ablation}
\centering
\begin{tabular}{c|c|c|c|ccccc}
\hline

 \multirow{2}*{}& \multicolumn{3}{c|}{Module}&\multirow{2}*{LFW}&\multirow{2}*{CALFW}& \multirow{2}*{CPLFW}&\multirow{2}*{AgeDB}&\multirow{2}*{Average}\\
\cline{2-4}
&RBM(w/o sg)&RBM&DBM&\\
\hline
\hline
\multirow{5}*{CASIA-Webface}& & & &98.8&91.0&85.4&90.2&91.35 \\

& & &\checkmark &99.2&92.0&87.3&91.9&92.6\\
&& \checkmark& &99.1&91.0&87.1&91.3&92.12\\

&\checkmark & & &98.7&90.6&85.4&90.3&91.25\\

&&\checkmark &\checkmark &99.3&92.5&87.6&92.1&92.88\\
\hline
\end{tabular}
\end{table*}


In all the experiments, the CNNs models are trained with stochastic
gradient descents (SGD). We set
the weight decay of 0.0005 and the momentum of 0.9.
The initial learning rate starts from 0.1 and is divided by 10 at the 5, 8, 11 epochs. The training process is finished at 15-th epoch. We set $\varepsilon=5.5$ and $\lambda=0.01$ in all the experiments.
The experiments are implemented by PyTorch \cite{paszke2017automatic} on NVIDIA Tesla V100 (32G).
We train the CNNs models from scratch and  only keep the feature embedding part without the final fully connected layer (512-D) during testing.

We use cosine distance to calculate the similarity. For the performance evaluation, we follow
the standard protocol of unrestricted with labeled outside data \cite{huang2008labeled} to report the performance on LFW, CALFW, CPLFW, AgeDB, RFW and AFW.
Considering the well solved on LFW, we further use the more challenging LFW BLUFR protocol to evaluate the proposed method.
On MegaFace, there are two challenges. We use large protocol in  Challenge 1 to evaluate
the performance of our approach. For the fair comparison, we also clean the
noisy images in Face Scrub and MegaFace by the noisy list \cite{deng2019arcface}.

\begin{figure}[tbp]
\includegraphics[scale=0.22]{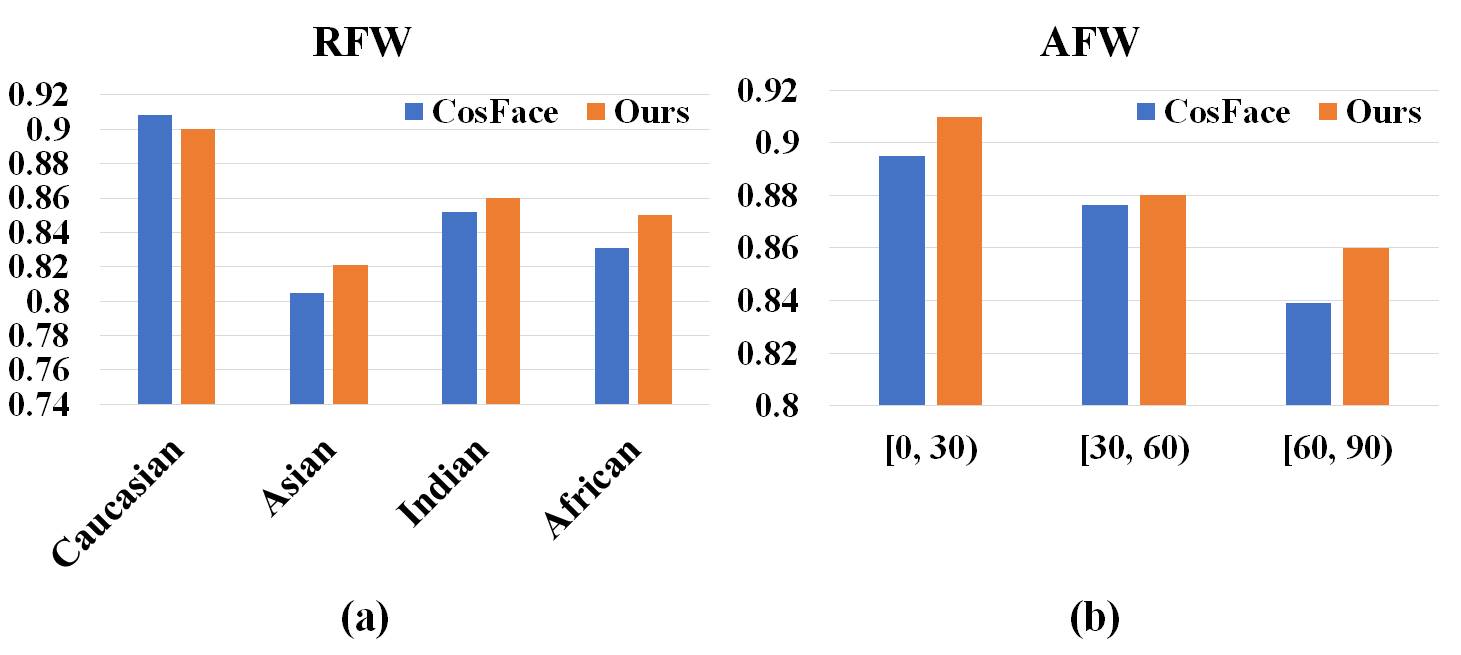}
\caption{\label{img:race-age-accuracy}(a) The performance on four testing subsets, Caucasian, Indian, Asian and
African in RFW. (b) The performance on three testing subsets, Young [0-30), Middle-aged [30-60) and Aged [60-90) in AFW.
}
\end{figure}

To the compared approaches, we compare the proposed method with the baseline Softmax loss and the recently popular
state-of-the-arts, including SphereFace \cite{liu2017sphereface}, CosFace \cite{wang2018cosface} and
 ArcFace \cite{deng2019arcface}.
\subsection{Ablation Study}
In this section, we investigate the effectiveness of each balancing module in the proposed method.

$\mathbf{Effectiveness \ \ of \ \ the \ \ RBM.}$ Recall that the RBM module consists of two main components: the residual enhancement and the soft gate. The soft gate produces a harmonizing coefficient to automatically control the magnitude of the residual attached to the top feature. When the soft gate is closed, i.e, $f(x)=1$ is constant for all samples, the RBM module degenerates to a conventional residual that loses the ability of distinguishing the head and tail domains.
From Table~\ref{tab:ablation}, we observe that
the combination of the residual enhancement and the soft gate brings large
improvements on all the datasets. The average performance of LFW, CALFW, CPLFW, AgeDB has been improved from 91.35 to 92.12. It is because RBM actually harmonizes the potential feature bias among different domains.

$\mathbf{Effectiveness \ \ of \ \ the \ \ Soft\ \ Gate.}$
The soft gate produces the coefficient DFI to control the magnitude of residual added on the original feature. In this experiment we analyze the effectiveness of the soft gate. As displayed in Table~\ref{tab:ablation}, the performance drops significantly without the soft gate. The average accuracy decreases 0.87$\%$. These results suggest that the improvement attained by the RBM block is not mainly due to the additional parameters, but its internal domain balancing mechanism.

$\mathbf{Effectiveness \ \ of \ \ the \ \ DBM.}$
We further validate the effectiveness of DBM that whether it can improve the poor generalization caused by the
long-tailed domain distribution. From the first row of each sub-boxes in Table~\ref{tab:ablation}, we can find that DBM boosts the
performance on all the datasets. The average performance is stably improved compared to the baselines,
presenting its contribution to mitigate the potential imbalance. Particularly, DBM achieves about $0.48\%$
average improvement over RBM, which indicates that balancing the contribution from different domains through loss function can better address the problem.

\subsection{Exploratory Experiments}
We first investigate how our method improves the performance on the different domains with different domain frequency.
We train Resnet50 on CASIA-Webface by CosFace and our method.
Figure~\ref{img:race-age-accuracy} shows the performances on different domains on two datasets. Firstly, for the Cosface, the accuracy of Caucasian on RFW is significantly higher than other races, and Asian gains the worse performance. Besides, on AFW, the Young subset acquires the highest accuracy while the performance on the aged persons degrades heavily. The performance decay confirms our thought that the popular methods is susceptible to the long-tailed domain distribution.
Secondly, our method consistently improves the performance on almost all the domains.
Particularly, the accuracy increases  more obviously on the tail domains, such as the Asian on RFW and $[60,90)$ aged persons on AFW, which indicates that the proposed method can alleviate the potential imbalance cross domains.

The nearest neighbor parameter $K$ in Eq.~\ref{eq:compactness1} plays an important role in DFI. In this part
we conduct an experiment to analyze the effect of $K$. We use CASIA-WebFace and ResNet18 to train the model with our method and evaluate
the performance on the LFW, CALFW and CPLFW as presented in Table~\ref{tab:explor}. We can conclude that the model without
DFI suffers from the poor performances on all these three benchmarks.
The model attains the worst result on all the datasets when $K=0$, where the model degenerates into the original form without balancing representation and margin supplied by RBM and DBM.
  The model obtains the highest accuracy at
$K=100$. However, when $K$ keeps increasing, the performances decrease to some extent because a too large K covers a too large region with sufficient samples and weakens the difference between head and tail domain.
\begin{table}[h]
\caption{Performance ($\%$) vs. $K$ on LFW, CALFW and CPLFW datasets, where $K$ is the number of nearest neighbor in Domain Frequency Indicator (DFI).}
\label{tab:explor}
\begin{center}
\begin{tabular}{|c||c|c|c|c|c|}
\hline
 K&0 & 100& 1,000&3,000&6,000\\
 \hline
 LFW& 98.8& 99.3&99.1&99.2&99.2\\
\hline
 CALFW& 91.0& 92.5&92.1&92.2&92.1\\
 \hline
 CPLFW& 85.4& 87.6&87.2&87.3&87.3\\
  \hline
\end{tabular}
\end{center}
\end{table}

\subsection{Evaluation Results}
\subsubsection{Results on LFW and LFW BLUFR}
LFW is the most widely used benchmark for unconstrained face recognition.
We use the common larget dataset MSIMV2 to train a ResNet50. Table~\ref{tab:lfw} displays the
the comparsion of all the methods on LFW testset. The proposed method improves the performance from $99.62\%$ to $99.78\%$.
Further, we evaluate our method on the more challenge LFW BLUFR protocol. The results are reported
in Table~\ref{tab:lfw_blufr}. Despite the limited improvement, our approach still achieves the best results compared to the state-of-the-arts.

\begin{table}[h]
\caption{Face verification ($\%$) on the LFW dataset. "Training Data" indicates the size of the training data involved. "Models" indicates the number of models used for evaluation.}
\label{tab:lfw}
\begin{center}
\begin{tabular}{|c||c|c|c|}
\hline
 Method &Training Data & Models& LFW\\
 \hline
Deep Face \cite{taigman2014closing} &4M & 3 & 97.35 \\
FaceNet \cite{schroff2015facenet}  &200M & 1 & 99.63 \\
DeepFR \cite{parkhi2015deep}&2.6M&1&98.95\\
DeepID2+ \cite{sun2015deeply}&300K&25&99.47\\
Center Face \cite{wen2016discriminative}&0.7M&1&99.28\\
Baidu \cite{liu2015targeting}&1.3M&1&99.13\\
\hline
Softmax & 5M & 1 &99.43  \\
SphereFace \cite{liu2017sphereface}&5M&1&99.57\\
CosFace \cite{wang2018cosface}&5M&1&99.62\\
ArcFace \cite{deng2019arcface}&5M&1&99.68\\
$\mathbf{Ours}$&5M&1&$\mathbf{99.78}$\\
\hline
\end{tabular}
\end{center}
\end{table}

\newcommand{\tabincell}[2]{\begin{tabular}{@{}#1@{}}#2\end{tabular}}
\begin{table}[h]
\caption{Face verification ($\%$) on LFW BLUFR protocol.}
\label{tab:lfw_blufr}
\begin{center}
\begin{tabular}{|c||c|c|}
\hline
 Method & \tabincell{c}{VR@FAR\\ =0.001\%}& \tabincell{c}{VR@FAR\\ =0.01\%} \\
 \hline
Softmax &87.53& 93.03   \\
SphereFace \cite{liu2017sphereface}  &98.50 & 99.17   \\
CosFace \cite{wang2018cosface}&98.70&99.20 \\
ArcFace \cite{deng2019arcface}&98.77&99.23 \\
$\mathbf{Ours}$&$\mathbf{98.91}$&$\mathbf{99.53}$ \\
\hline
\end{tabular}
\end{center}
\end{table}

\subsubsection{Results on CALFW, CPLFW and AgeDB}
Table \ref{tab:lfw_calfw} shows the performances on  CALFW, CPLFW and AgeDB, respectively.
We also use MSIMV2 to train the ResNet50. The results show the similar
treads that emerged on the previous test sets. Particularly, the margin-based methods attain better results
than the simple softmax loss for face recognition. Our proposed method, containing efficient domain balancing mechanism, outperforms all the other methods on these
three datasets. Specifically, our method achieves $95.54\%$ average accuracy, about $0.4\%$ average improvement
over ArcFace.
\begin{table}[h]
\caption{Face verification ($\%$) on CALFW, CPLFW and AgeDB.}
\label{tab:lfw_calfw}
\begin{center}
\begin{tabular}{|c|c|c|c|}
\hline
 Method &CALFW& CPLFW&AgeDB \\
 \hline
Softmax &89.41& 81.13& 94.77 \\
SphereFace \cite{liu2017sphereface} &90.30 & 81.40& 97.30   \\
CosFace \cite{wang2018cosface} &93.28&92.06& 97.70\\
ArcFace \cite{deng2019arcface} &95.45&92.08& 97.83\\
$\mathbf{Ours}$&$\mathbf{96.08}$&$\mathbf{92.63}$& $\mathbf{97.90}$\\
\hline
\end{tabular}
\end{center}
\end{table}
\subsubsection{Results on  MegaFace}
We also evaluate our method on the large Megaface testset. Table \ref{tab:megaface} displays the identification and verification performances. In particular, the proposed method surpasses the best approach
ArcFace by an obvious margin (about $0.82\%$ at Rank-1 identification rate and $0.68\%$ verification rate).
The reason behind may be  that the proposed balancing strategy can efficiently mitigate the potential
impact of the long-tailed domain distribution, which is ubiquitous in the real-world application.

\begin{table}[h]
\caption{Face identification and verification on MegaFace Challenge1.
"Rank 1"  refers to the rank-1 face identification accuracy, and "Ver" refers to the
face verification TAR at $10^{-6}$ FAR.}
\label{tab:megaface}
\begin{center}
\begin{tabular}{|c||c|c|}
\hline
 Method &Rank1 (\%) & Ver (\%) \\
 \hline
DeepSense V2 &81.29& 95.99   \\
YouTu Lab  &83.29 & 91.34   \\
Vocord-deepVo V3&91.76&94.96 \\
 \hline
SphereFace \cite{liu2017sphereface}&92.05&92.42 \\
CosFace \cite{wang2018cosface}&94.84&95.12 \\
ArcFace \cite{deng2019arcface}&95.53&95.88 \\
$\mathbf{Ours}$&$\mathbf{96.35}$&$\mathbf{96.56 }$ \\
\hline
\end{tabular}
\end{center}
\end{table}


\section{Conclusion}
In this paper, we investigate a novel long-tailed domain problem in the real-world face recognition, which refers to few common domains and many more rare domains. A novel Domain Balancing mechanism is proposed to deal with this problem, which contains three  components, Domain Frequency Indicator (DFI), Residual Balancing Mapping (RBM) and Domain Balancing Margin (DBM). Specifically, DFI is employed to judge  whether a class belongs to a head domain or a tail domain. RBM introduces a light-weighted residual controlled by the soft gate. DBM assigns an adaptive
margin to balance the contribution from different domains. Extensive analyses and experiments on several face recognition benchmarks demonstrate that the proposed method can effectively enhance the discrimination and achieve superior accuracy.

\section*{Acknowledgement}
This work has been partially supported by the Chinese National Natural Science Foundation Projects $\#61876178, \#61806196, \#61976229, \#61872367$

{\small
\bibliographystyle{ieee_fullname}
\bibliography{egbib}
}

\end{document}